\pdfoutput=1

\documentclass[11pt]{article}

\usepackage[final]{acl}

\usepackage{times}
\usepackage{latexsym}
\usepackage{booktabs}
\usepackage[T1]{fontenc}

\usepackage[utf8]{inputenc}

\usepackage{microtype}

\usepackage{inconsolata}

\usepackage{graphicx}
\usepackage{soul}
\usepackage{amsmath} 
\title{To Preserve or To Compress: An In-Depth Study of Connector Selection in Multimodal Large Language Models}

\usepackage{booktabs}
\usepackage{graphicx}
\usepackage{float} 
\usepackage{multirow}
\usepackage{array}
\usepackage{makecell}
\usepackage{amssymb}

\author{Junyan Lin\textsuperscript{1}, Haoran Chen\textsuperscript{1}, Dawei Zhu\textsuperscript{2}, Xiaoyu Shen\textsuperscript{1}\thanks{Corresponding Author}\\
\textsuperscript{1}Digital Twin Institute, Eastern Institute of Technology, Ningbo, China, \\
\textsuperscript{2}Saarland University, Saarland Informatics Campus\\
  \texttt{linjyan00@gmail.com \qquad xyshen@eitech.edu.cn}} 
  
\begin{document}

\maketitle

\begin{abstract}
In recent years, multimodal large language models (MLLMs) have garnered significant attention from both industry and academia. However, there is still considerable debate on constructing MLLM architectures, particularly regarding the selection of appropriate connectors for perception tasks of varying granularities. This paper systematically investigates the impact of connectors on MLLM performance. Specifically, we classify connectors into feature-preserving and feature-compressing types. Utilizing a unified classification standard, we categorize sub-tasks from three comprehensive benchmarks, MMBench, MME, and SEED-Bench, into three task types: coarse-grained perception, fine-grained perception, and reasoning, and evaluate the performance. Our findings reveal that feature-preserving connectors excel in \emph{fine-grained perception} tasks due to their ability to retain detailed visual information. In contrast, feature-compressing connectors, while less effective in fine-grained perception tasks, offer significant speed advantages and perform comparably in \emph{coarse-grained perception} and \emph{reasoning} tasks. These insights are crucial for guiding MLLM architecture design and advancing the optimization of MLLM architectures. \footnote{Our code is available at \url{https://github.com/EIT-NLP/Connector-Selection-for-MLLM}.}
\end{abstract}

\section{Introduction}

\begin{figure*}[ht]
\includegraphics[width=\textwidth]{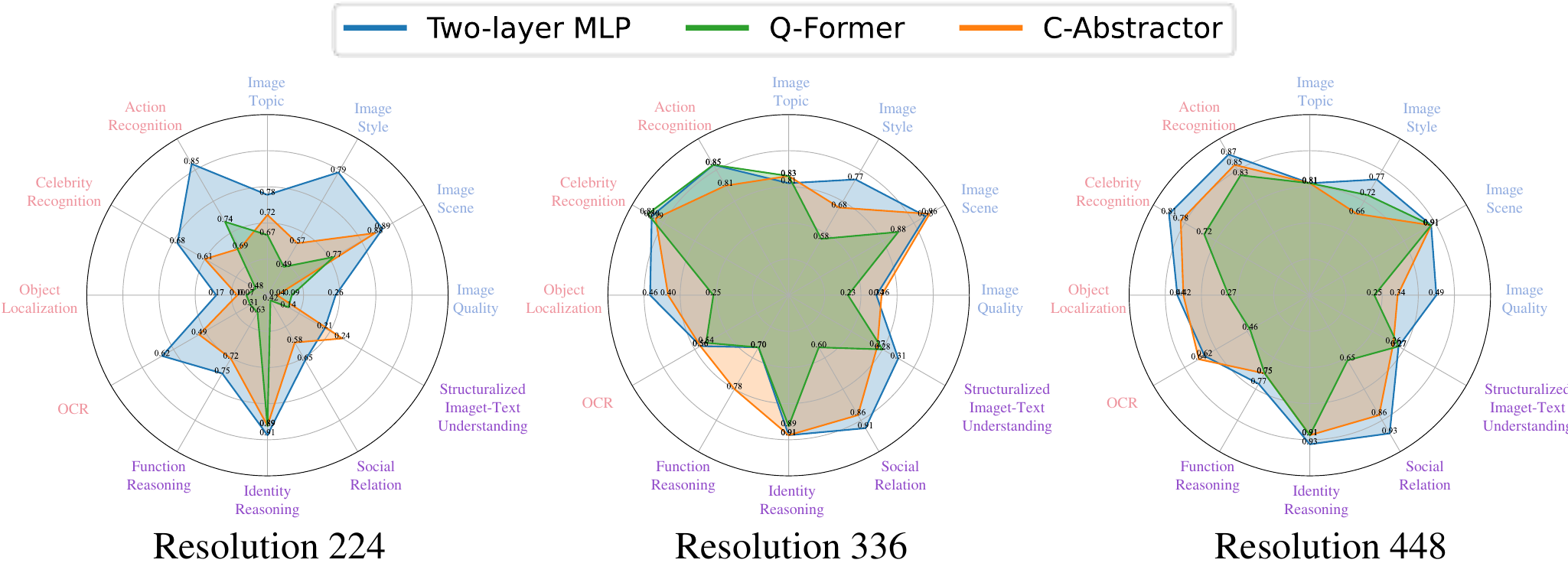}
\caption{Comparison of radar chart performance at 224, 336, and 448 resolutions across coarse-grained perception, fine-grained perception, and reasoning tasks on MMBench. Each task includes four sub-tasks: Image Quality, Image Scene, Image Style, and Image Topic for coarse-grained perception; Action Recognition, Celebrity Recognition, Object Localization, and OCR for fine-grained perception; and Function Reasoning, Identity Reasoning, Social Relation, and Structuralized Image-Text Understanding for reasoning tasks.}
\label{fig:radar_charts}
\end{figure*}

Large language models (LLMs) have made significant advances in recent years, demonstrating remarkable capabilities in understanding and generating text~\citep{brown2020language,su2022welm,jiang2023mistral,bai2023qwen,touvron2023llama,zhang2024impact,su2024unraveling}.
Recently multimodal large language models (MLLMs) have emerged as a hot topic in both academia and industry due to their potential to handle multiple modalities, such as text and vision, in a unified framework~\citep{wang2023cogvlm,alayrac2022flamingo,gao2024sphinx}. However, training a unified architecture from scratch across different modalities is often resource-intensive and time-consuming, making it feasible only for a limited number of large companies with substantial computational resources~\cite{team2024chameleon,zhou2024transfusion}.  As a result, researchers commonly adopt a \textbf{connector}-based approach, which fully leverages the existing powerful capabilities of a pre-trained language model~\cite{li2023blip,liu2024visual}. This connector bridges the gap by transforming visual information from the encoder into vector representations that the LLM can process and understand. Through this method, the pre-trained text-based LLM is empowered to perceive and interpret visual data, enabling it to perform a wide range of visual tasks without requiring a complete retraining of the model from scratch~\cite{yin2023survey}.

Designing an optimal MLLM architecture remains a crucial and intriguing research area. While prior studies \citep{karamcheti2024prismatic,mckinzie2024mm1,laurenccon2024matters} have investigated various factors affecting MLLM performance, a significant gap persists in the detailed examination of the key component: \textbf{the connector}.

We categorize connectors into two types: \textbf{feature-preserving connectors}, which retain visual feature details by maintaining patch numbers, and \textbf{feature-compressing connectors}, which reduce computational load by abstracting patches into a specified number. Different studies have conflicting views: \citet{lin2023sphinx,chen2024lion} contends that feature-compressing connectors are suitable for coarse-grained perception tasks but perform weakly on fine-grained perception tasks. In contrast, \citet{mckinzie2024mm1} observes little difference between the two. Although these studies provide experimental evidence, further exploration is needed to understand how connectors perform across varying perception granularities.

To address this gap, this paper aims to meticulously investigate the effects of various connectors on tasks of different perception granularities. Building on the construction guidelines of MMBench \citep{liu2023mmbench}, we evaluate the impact of connectors on the performance of MLLMs across three task types: coarse-grained perception, fine-grained perception, and reasoning. Our extensive experiments thoroughly explore connector performance across these varying perception granularities. Several noteworthy conclusions are drawn from testing on three multitask benchmarks. Figure~\ref{fig:radar_charts} shows the performance of different connectors on the three tasks as well as the corresponding sub-tasks. Our main contributions are summarized as follows:

\begin{enumerate}
\item We conduct a comprehensive analysis of different connectors from multiple perspectives, including loss curves, compressed token number, image resolution, and performance metrics across tasks of different granularities.
\item We demonstrate that feature-compressing connectors significantly underperform in fine-grained perception tasks compared to feature-preserving connectors, while maintaining comparable performance in coarse-grained perception tasks.
\item We analyze the impact of different pooling methods within feature-compressing connectors, revealing that simpler pooling methods generally lead to more effective training and better overall performance.
\end{enumerate}

\section{Related Work}
  
Connectors play a crucial role in aligning multimodal data within MLLMs, with various types. Based on whether the patch number of visual features is retained or reduced, we classify connectors into two categories: feature-preserving connectors and feature-compressing connectors.LLaVA \citep{liu2024visual} employs a single-layer linear projection as its connector, whereas LLaVA-1.5 \citep{liu2024improved} enhances this design by adding a GELU activation function and an extra linear projection. These feature-preserving connectors are designed to retain the details of visual features. Emu2 \citep{sun2024generative} uses a local average pooling strategy to standardize visual features into a uniform number of patches. BLIP-2 \citep{li2023blip} utilizes the Q-Former, a cross-attention connector that uses a fixed number of learnable queries to interact with visual features, enabling global weighted pooling. HoneyBee \citep{cha2024honeybee} introduces the C-Abstractor, which utilizes convolutional neural networks to perform local weighted pooling based on Emu2's local average pooling, effectively extracting local features. These feature-compressing connectors adjust the length of feature patch number, thereby optimizing computational resources while reserving key information.

To explore the impact of various components and parameters in MLLMs on performance, numerous studies have been conducted. \citet{karamcheti2024prismatic} rigorously investigates MLLMs along key design axes, such as optimization procedures, image processing, pretrained visual representations, language models, and scaling properties. MM1 \citep{mckinzie2024mm1} aim to identify important design principles and lessons for constructing MLLMs through comprehensive ablation studies on architecture components, data choices, and training procedures. Additionally, Idefics2 \citep{laurenccon2024matters} conducts extensive experiments around pre-trained models, architecture choice, data, and training methods to bring experimental clarity to core design choices in building MLLMs.

Their findings offer useful initial insights but require a more detailed analysis of task-specific performance for greater depth and applicability. Specifically, they lack a comprehensive examination of tasks with varying granularities, such as coarse-grained perception, fine-grained perception, and reasoning \citep{liu2023mmbench}.

To address this, our paper thoroughly investigates the effects of various connectors on MLLMs, focusing on their performance across the aforementioned tasks. This comprehensive analysis aims to deepen our understanding of connectors' impact and guide targeted connector selection in future model design stages based on specific tasks.

\begin{figure*}[t]
  \includegraphics[width=2\columnwidth]{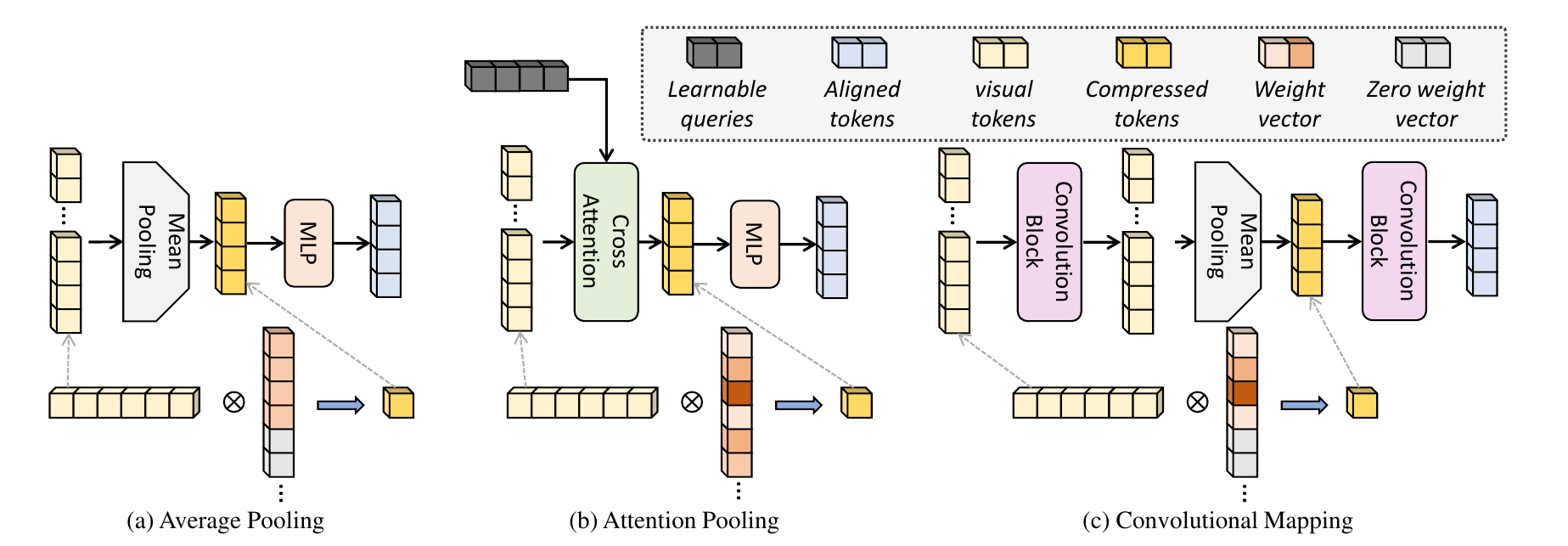}
  \caption{The structure of different visual-language connectors. The upper part of the figure shows the overall structure of various connectors, while the lower part provides a simplified visualization during compression. (a) The Average Pooling-based connector compresses features by averaging visual tokens within local windows (b) The Attention Pooling-based connector uses cross-attention between learnable queries and visual tokens to abstract visual tokens into a certain number of compressed tokens. Each compressed token is derived from all visual tokens with weighted contributions. (c) The Convolutional Mapping-based connector uses convolution operations to enhances local context modeling while reducing the number of tokens. Each compressed token is derived from the visual tokens within local windows with weighted contributions.}
  \label{fig:connectors}
\end{figure*}

\section{Taxonomy of Connectors}

\subsection{Preliminaries}

Multimodal large language models (MLLMs) generally consist of three key components: a visual encoder $E$, a connector $C$, and an LLM. For a given visual input $V$, the encoder $E$ extracts visual features $f \in \mathbb{R}^{P \times d_v}$, where $P$ is the number of visual patches and $d_v$ is the channel dimension. The connector $C$, which is a crucial component, then aligns these visual features with the word embedding space as follows:
\begin{equation}
  \begin{aligned}
  C: \mathbb{R}^{P \times d_v} &\rightarrow \mathbb{R}^{Q \times D} \\
  x &= C(f)
  \end{aligned}
\end{equation}
Here, $x \in \mathbb{R}^{Q \times D}$ represents the visual tokens that are input into the LLM, where $Q$ is the number of visual tokens and $D$ is the hidden size of the LLM. We categorize connectors into two types: feature-preserving connectors, where $P = Q$, and feature-compressing connectors, where $P > Q$.

\subsection{Feature-Preserving Connector}
Feature-preserving connectors maintain the patch number of visual features (i.e., $P = Q$) and are typically composed of components such as linear layers and activation layers.  While they can retain detailed information, the computational complexity of the model grows exponentially with the visual token number. Existing feature-preserving connectors can be classified into linear and nonlinear types based on whether they include nonlinear operations. For example, the connector in LLaVA \citep{liu2024visual} can be classified into linear type because it only contains a linear layer, as shown below:
\begin{equation}
x = W f
\end{equation}
where $W \in \mathbb{R}^{d_v \times D}$ is a trainable projection matrix, that maps the visual features $f \in \mathbb{R}^{P \times d_v}$ to the word embedding space. The connector used in LLaVA-1.5 \citep{liu2024improved} can be classified into the nonlinear type because it incorporates an activation function and an additional linear layer on top of the basic linear type, as shown below:
\begin{equation}
x = W^{(2)}  \phi(W^{(1)} f)
\end{equation}
where $W^{(1)} \in \mathbb{R}^{d_v \times D}$ and $W^{(2)} \in \mathbb{R}^{D \times D}$ are trainable projection matrices, and $\phi$ denotes a non-linear activation function GELU. The inclusion of the activation function allows the nonlinear connectors to capture more intricate patterns and dependencies in the data, enhancing the model's ability to manage complex visual-textual relationships.

\subsection{Feature-Compressing Connector}

Feature-compressing connectors reduce the number of visual tokens $Q$ through various strategies, aiming to preserve visual information while optimizing computational efficiency. Based on MM1 \citep{mckinzie2024mm1}, we categorize feature-compressing connectors into three types: average pooling, attention pooling, and convolutional mapping, as shown in Figure~~\ref{fig:connectors}. They generally operate in two steps. The first step involves using a pooling operation $\mathcal{P}$ to reduce the patch number $P$ of visual feature $f$ to $Q$ ($Q < P$) as follows:
\begin{equation}
f' = \mathcal{P}(f)
\end{equation}
where $f' \in \mathbb{R}^{Q \times d_v}$ represents the compressed visual features. The second step is consistent with the feature-preserving connector, where the compressed visual features $f'_v$ are projected into the word embedding space using a transformation $\mathcal{T}$:
\begin{equation}
x = \mathcal{T}(f')
\end{equation}
where $\mathcal{T}$ can be a multi-layer perception (MLP) or convolutional layer that maps $f'$ to $x \in \mathbb{R}^{Q \times D}$.

\paragraph{Average Pooling}
This type of connector uses average pooling as $\mathcal{P}$ to reduce the number of tokens. For a given set of $n$ feature patches in $f$, the average pooling operation can be formulated as follows:
\begin{equation}
f'_i = \frac{1}{n} \sum_{j=1}^{n} f_{(i-1)n+j}
\end{equation}
where $f'_{v,i}$ represents the $i$-th averaged feature patch in $f'$ and $f_{(i-1)n+j}$ represents the $j$-th feature patch in the $i$-th group of $f$. After obtaining the compressed visual features $f'$, we directly apply the connector from LLaVA-1.5 as the transformation $\mathcal{T}$ to project $f'$ into the word embedding space.

\paragraph{Attention Pooling}
This type of connector uses cross-attention as $\mathcal{P}$ to reduce the number of tokens. The patch number $P$ is compressed by performing cross-attention between a set of learnable queries $Q \in \mathbb{R}^{Q \times d_c}$ and the visual features $f$, resulting in $f' \in \mathbb{R}^{Q \times d_c}$, where $d_c$ is the hidden size of the cross-attention. The cross-attention mechanism can be formulated as follows:
\begin{equation}
  \begin{aligned}
    K         &= W_k f, \quad V = W_v f \\
    A         &= \text{softmax}\left(\frac{QK^\top}{\sqrt{d_c}}\right) \\
    f'_i  &= \sum_{j=1}^{P} A_{ij} V_j
  \end{aligned}
\end{equation}
where $K, V \in \mathbb{R}^{P \times d_c}$ are the key and value matrices obtained by projecting the visual features $f$ using the projection matrices $W_k, W_v \in \mathbb{R}^{d_v \times d_c}$, respectively. $A \in \mathbb{R}^{Q \times P}$ is the attention weight matrix, and $A_{ij}$ represents the attention weight between the $i$-th query and the $j$-th visual feature. The compressed visual feature $f'_i$ is obtained by weighted summation of the value vectors $V_j$. After obtaining the compressed visual features $f'$, the transformation $\mathcal{T}$ is consistent with the approach used in average pooling connector.

\paragraph{Convolutional Mapping}
This type of connector uses a combination of convolutional layers and average pooling as $\mathcal{P}$ to reduce the number of tokens. The patch number $P$ is compressed by first applying convolutional layers followed by average pooling, resulting in $f' \in \mathbb{R}^{Q \times d_v}$, where $d_v$ is the channel dimension of the visual features. The transformation $\mathcal{T}$ is then applied using additional convolutional layers to project the compressed visual features into the word embedding space. The overall process can be formulated as follows:

\begin{equation}
  \begin{aligned}
    f'_i &= \frac{1}{n} \sum_{j=1}^{n} \left( W_j \cdot f_{(i-1)n+j} \right) \\
    x_i  &= \sum_{k=-K}^{K} W'_k \cdot f'_{i+k}
  \end{aligned}
\end{equation}
where $x_i$ represents the $i$-th projected visual token, obtained by applying convolutional layers represented by weights $W'_k$ over the compressed features $f'_{i+k}$. For simplification, the initial convolutional layers and average pooling are represented as a local weighted average, denoted by $W_j$.

\paragraph{Characteristics}

Average pooling is a simple and efficient feature-compressing connector that quickly reduces patch numbers without adding any parameters, making it easy to train. However, it may lead to the loss of local information, reducing its effectiveness in fine-grained perception tasks. Attention pooling, utilizing a global weighted mechanism, retains more global information and offers a higher theoretical performance limit. Despite this, it has higher computational complexity and the most additional parameters due to the learnable queries and projection matrices, making it the most challenging to train. Furthermore, it may struggle with fine-grained tasks because the attention mechanism can find it difficult to preserve local image information \citep{dosovitskiy2020image,park2022vision}. Convolutional mapping effectively preserves local details and involves a moderate number of parameters, striking a balance between parameter efficiency and the ability to capture fine-grained details. However, it lacks the capability to intuitively capture global features. These characteristics are intuitive, but their actual effectiveness and trade-offs need to be empirically validated through extensive experiments across tasks with varying perception granularities.

\section{Experimental Settings}
\label{sec:setting}
In this section, we present the experimental settings employed in our study, including the criteria for perception granularity (\ref{sec:granularity}), the benchmarks utilized for evaluation (\ref{sec:benchmarks}), and the specific implementation details of our models (\ref{sec:details}). Each aspect is carefully detailed to ensure a comprehensive understanding of our experiment.

\subsection{Perception Granularity}
\label{sec:granularity}
The partition criterion for coarse-grained and fine-grained perception vary across different benchmarks. For example, MMBench categorizes tasks like `Image Style' and `Image Scene' under coarse-grained perception, which focuses on the global attributes and overall context of the image, while tasks like `Object Localization' fall under fine-grained perception, focusing on the local details and specific features within the image. MME also differentiates between coarse-grained perception and fine-grained perception tasks, but its criteria focus more on testing the knowledge resources of MLLM, rather than the perspective of spatial scope.

\begin{figure}[ht]
    \centering
  \includegraphics[width=0.90\columnwidth]{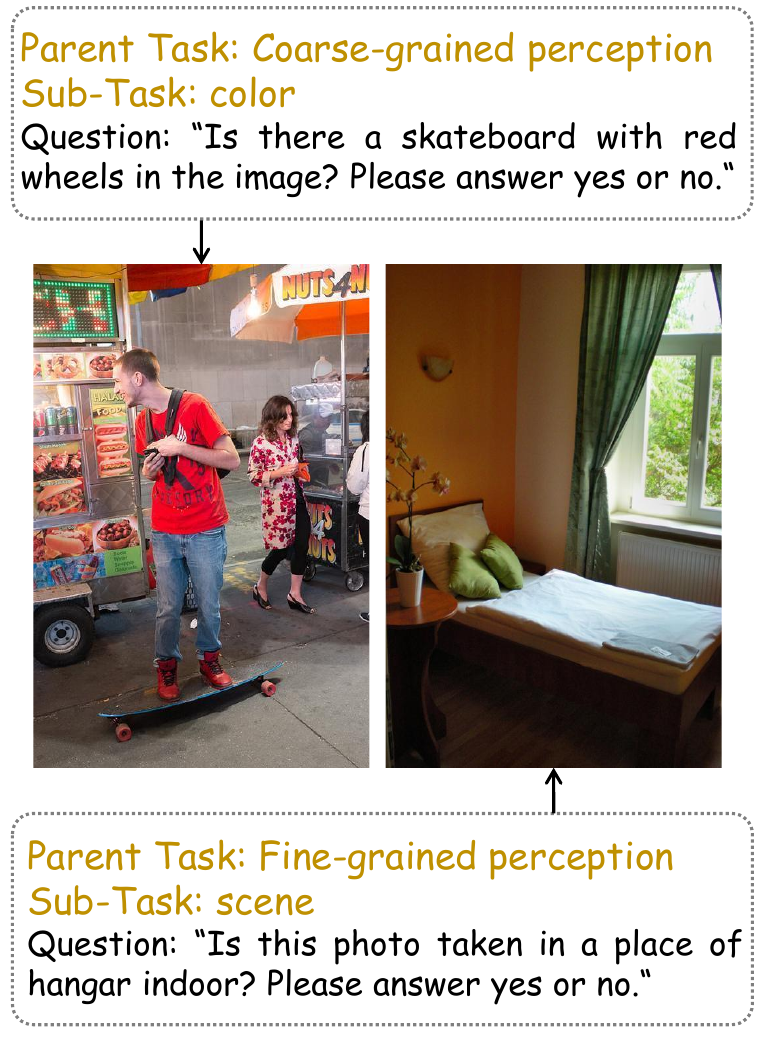}
  \caption{
Examples of conflicting partition criterion for perception granularity in the MME benchmark.
  }
  \label{fig:conflict_subtask}
\end{figure}

For instance, the task of `Object Localization' in MME is considered a coarse-grained perception task and `Scene Recognition' is classified as a fine-grained perception task. However, in MMBench, they will be divided into coarse-grained perception task and fine-grained perception task, respectively. Figure~\ref{fig:conflict_subtask} further illustrates this discrepancy. The left image is selected from the `Color' sub-task, which is categorized as a coarse-grained perception task in MME. However, it actually focuses on local image details, which would reclassify it as a fine-grained perception task in MMBench. Conversely, the right image is selected from the `Scene' sub-task, which is categorized as a fine-grained perception task in MME, but it actually focuses on the overall context of the image, making it a coarse-grained perception task in MMBench.

Many methods exhibit conflicting views on the ability of different connectors to handle different granularities \citep{lin2023sphinx,chen2024lion, mckinzie2024mm1}. To explore this issue, based on MMBench, we define coarse-grained perception as the ability to perceive image-level features, such as overall concepts and context. In contrast, fine-grained perception refers to object-level details within the image, such as identifying specific features of individual objects. By analyzing performance in coarse-grained and fine-grained perception tasks, we can determine whether feature-preserving connectors and feature-compressing connectors excel in specific perception tasks. Additionally, by examining reasoning tasks, we can more accurately assess the impact of different types of connectors on the model's ability to understand and integrate multimodal information.

\subsection{Benchmarks}
\label{sec:benchmarks}
To explore and evaluate various types of connectors, we utilize three well-established benchmarks with sub-task labels: MMBench \citep{liu2023mmbench}, MME \citep{fu2023mme}, and SEED-Bench \citep{li2023seedbench}. We reference the coarse-grained and fine-grained perception tasks defined above to reclassify the sub-tasks of MME and SEED-Bench. Detailed information on the sub-tasks reclassification can be found in Table~\ref{tab:reclassification_seedbench}, Table~\ref{tab:reclassification_mme}, and Table~\ref{tab:all_parenttask} in Appendix~\ref{sec:reclass}.

\subsection{Implementation Details} 
\label{sec:details}
Given the focus of this paper on comparing connectors, we largely adhere to the configuration of LLaVA-1.5, with exceptions for connector modifications and using LLaMA 2 \citep{touvron2023llama} as the LLM. The visual encoder utilized is CLIP ViT-L/14 \citep{radford2021learning} with resolutions of 224 and 336. We keep the learning rate, batch size, training phases, and data usage consistent with LLaVA-1.5. For images with a resolution of 448, we refer to MM1 and employ position embedding interpolation to adapt CLIP ViT-L/14 from a resolution of 336 to 448. Considering that LoRA-based LLaVA-1.5 performs on par with the fully fine-tuning setting across the three benchmarks, we opt for the LoRA approach \citep{hu2021lora} to save computational resources. Refer to Table~\ref{tab:training_config} in Appendix~\ref{sec:exp_details} for detailed connector configurations.

\section{Results}
\label{sec:results}

To verify the ability of different connectors to perceive different image granularities and assess reasoning capabilities, we evaluate the performance of feature-preserving and feature-compressing connectors with different image resolutions across three key tasks: coarse-grained perception, fine-grained perception, and reasoning.

\subsection{Effects of Feature-Preserving Connector}

\begin{figure}[htbp]
    \centering
    \includegraphics[width=1\columnwidth]{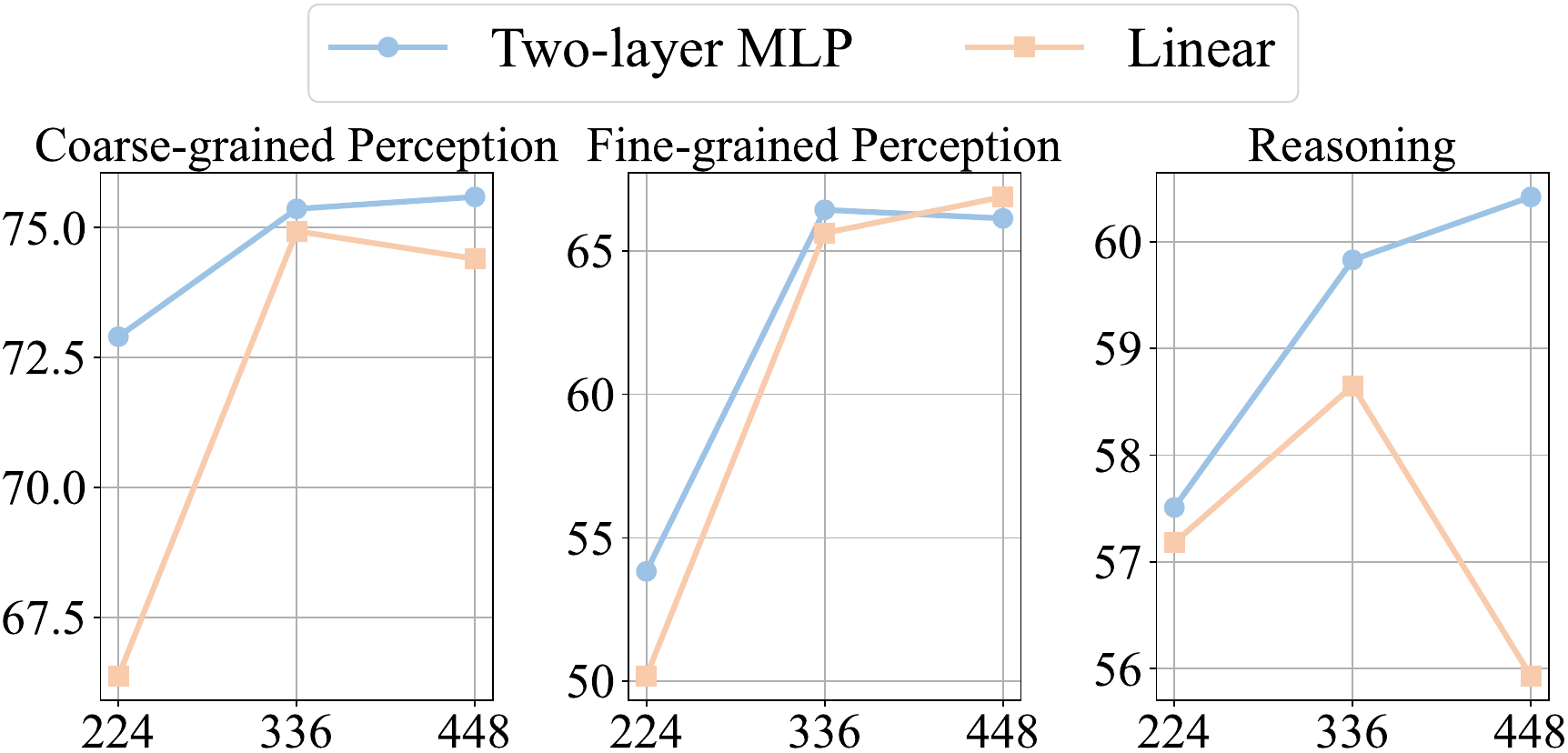}

    \caption{Comparison of two-layer MLP and linear connectors on coarse-grained, fine-grained perception, and reasoning tasks at resolutions of 224, 336, and 448.}
\label{fig:mlpvslinear}    
\end{figure}

To assess the performance of feature-preserving connectors, we compare the linear type (referred to as the linear connector) with the nonlinear type (referred to as the two-layer MLP connector) across three tasks: coarse-grained perception, fine-grained perception, and reasoning. As shown in Figure~\ref{fig:mlpvslinear}, using two-layer MLP connector consistently outperform the linear connector in all task groups at a resolution below 448. 

When the resolution is increased to 448, although the linear connector performs on par with the two-layer MLP connector in fine-grained perception, it suffers a substantial performance drop in other tasks, particularly in reasoning. We hypothesize that a linear mapping may struggle balancing both perception and reasoning at high resolution. In contrast, the reasoning ability is further enhanced when using two-layer MLP with higher resolution.

\subsection{Impact of Compressed Token Number}

\begin{figure}[htbp]
    \centering
    \includegraphics[width=1\linewidth]{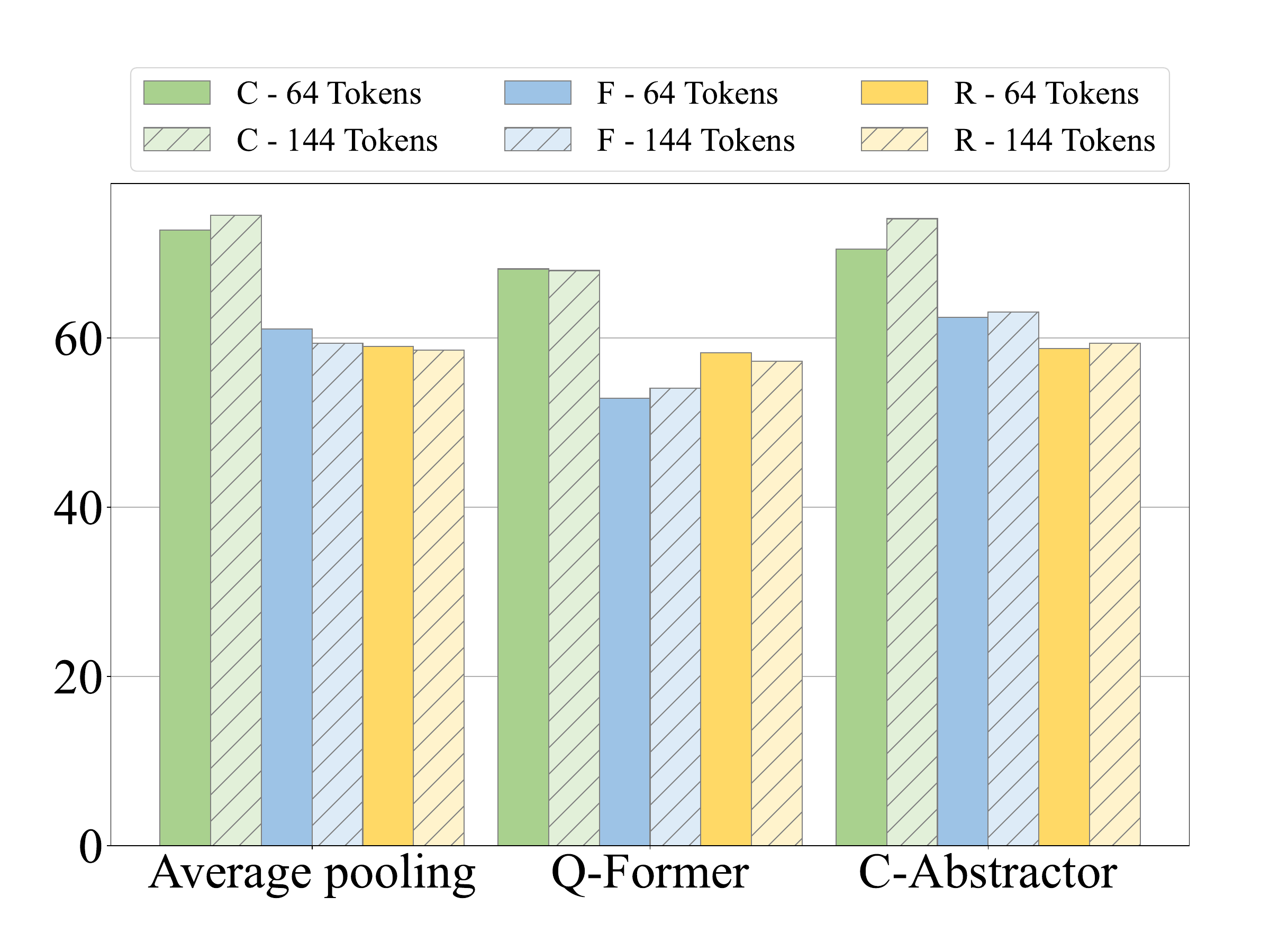}
    \caption{Analysis of the impact of different compressed token numbers on the performance of coarse-grained perception (C), fine-grained perception (F), and reasoning (R) tasks.}
    \label{fig:different_tokens}
\end{figure}

The compressed token number $Q$ is an important parameter. We compare two widely used values: 64 and 144. We fix the resolution at 336 and evaluate average pooling, Q-Former, and C-Abstractor across three tasks. The results are shown in Figure~\ref{fig:different_tokens}. It can be seen that while 144 tokens generally provide a slight improvement in performance over 64 tokens, the difference is not substantial, indicating that both 64 and 144 tokens are adequate for robust image information extraction.

Additionally, to further demonstrate the impact of the compressed token number, we present the loss curves of different feature-compressing connectors with different compressed token numbers during the pretraining and finetuning stages in Figure~\ref{fig:loss_compare} in Appendix~\ref{sec:loss_curves}. It can be observed that the loss curves from 144 tokens show marginally better convergence than those from 64 tokens, especially for C-Abstractor. Considering this slight performance gain, we ultimately choose 144 tokens as the standard setting for feature-compressing connectors for subsequent comparisons.

\subsection{Impact of Image Resolution} 
We further explore the effect of image resolution on the metric performance of feature-preserving and feature-compressing connectors. Detailed experiments and analyses, as shown in Table~\ref{tab:resolution_increase}, reveal that increasing the resolution from 224 to 336 enhances performance across all connector types for the three tasks, with the most significant improvements observed in fine-grained perception tasks, followed by coarse-grained perception tasks, and the least improvement in reasoning tasks. However, further increasing the resolution from 336 to 448 yields only marginal performance gains. Specifically, for feature-preserving connectors, the resolution increase from 224 to 336 results in improvements of 12.6\% in fine-grained perception, 2.5\% in coarse-grained perception, and 2.3\% in reasoning tasks. For feature-compressing connectors, the improvements are 13.9\%, 9.2\%, and 4.3\%, respectively. When the resolution is increased from 336 to 448, the performance changes for the former are 2.5\%, 0.2\%, and 0.6\%, while for the latter, the changes are -0.5\%, -1.0\%, and 0.9\%. We attribute this to the diminishing returns of higher resolutions and the current insufficient training data to support them.

\begin{table}[ht]
\centering

\resizebox{\columnwidth}{!}{
\begin{tabular}{l>{\centering\arraybackslash}m{2cm} >{\centering\arraybackslash}m{2cm} >{\centering\arraybackslash}m{2cm}}
\toprule
\textbf{Connectors} & $C$ & $F$ & $R$ \\
\midrule
\textbf{Two-layer MLP} & & & \\
\quad 224 & 72.9  & 53.83 & 57.51 \\
\quad 336 & 75.36 & 66.43 & 59.83 \\
\quad 448 & 75.59 & 66.13 & 60.42 \\
\midrule
\textbf{Average pooling} & & & \\
\quad 224 & 70.11 & 53.47 & 57.56 \\
\quad 336 & 74.54 & 59.37 & 58.59 \\
\quad 448 & 74.29 & 64.84 & 60.53 \\
\midrule
\textbf{Q-Former} & & & \\
\quad 224 & 59.29 & 42.71 & 51.97 \\
\quad 336 & 67.99 & 54.09 & 57.29 \\
    \quad 448 & 69.62 & 52.83 & 56.99 \\
\midrule
\textbf{C-Abstractor} & & &  \\
\quad 224 & 64.93 & 49.33 & 55.05 \\
\quad 336 & 74.12 & 63.11 & 59.39 \\
\quad 448 & 73.05 & 62.62 & 60.31 \\
\midrule

\end{tabular}}
\caption{Comparison of two-layer MLP, average pooling with 144 tokens, C-Abstractor with 144 tokens, and Q-Former with 144 tokens on coarse-grained perception (C), fine-grained perception (F), and reasoning (R) tasks.}
\label{tab:resolution_increase} 
\end{table}

Figure~\ref{fig:loss_compare} in Appendix~\ref{sec:loss_curves} clearly illustrates the loss curves of all connectors at different resolutions. It can be seen that in most cases, increasing the resolution from 224 $\times$ 224 to 336 $\times$ 336 generally results in a decrease in training loss. However, when the resolution is further increased from 336 $\times$ 336 to 448 $\times$ 448, only the fine-tuning loss of the two-layer MLP decreases, while the others either remain unchanged or increase. This observation is consistent with the evaluation metrics.

\begin{table}[htbp]
\centering
\resizebox{\columnwidth}{!}{
\begin{tabular}{l>{\centering\arraybackslash}m{2cm} >{\centering\arraybackslash}m{2cm} >{\centering\arraybackslash}m{2cm}}
\toprule
\textbf{Connectors} & $C$ & $F$ & $R$ \\
\midrule
Average pooling & 74.29 & 64.84 & 60.53 \\
Q-Former & 69.62 & 52.83 & 56.99 \\
C-Abstractor & 73.05 & 62.62 & 60.31 \\

\bottomrule
\end{tabular}
}
\caption{Comparison of feature-compressing connectors at 448 resolution on coarse-grained perception (C), fine-grained perception (F), and reasoning (R) tasks, each with the same compressed token number 144.}
\label{tab:compress_type} 
\end{table}

\subsection{Effects of Feature-Compressing Connector}

To explore the impact of different feature-compressing connectors on model performance, we conduct a detailed comparison on the three tasks under the settings of 448$\times$448 resolution and 144 compressed token number, as shown in Table~\ref{tab:compress_type}. Overall, the performance of average pooling and C-Abstractor is similar, while Q-Former performs significantly worse. Specifically, for coarse-grained perception tasks, Q-Former does not show as large a performance gap compare to other connectors as it does in fine-grained perception tasks. This might be because, the self-attention mechanism disrupts the original positional information which is important in fine-grained perception tasks. However, this does not fully explain why Q-Former also performs poorly in coarse-grained perception tasks. To explain this phenomenon, we present the training loss curves at Figure~\ref{fig:loss_compare} in Appendix~\ref{sec:loss_curves}. The curves show that Q-Former's loss decreases more slowly than the losses from other connectors. This indicates that Q-Former is more challenging to train, likely due to insufficient training data to support such a complex mechanism.

In summary, simple average pooling suffices for most tasks as LLMs can implicitly extract image information from visual tokens. Extensive interference in token extraction at the projector stage is not necessary. Complex connectors like Q-Former may require more aligned data for better results.

\section{Suggestions for Connector Selection}
\label{sec:5}

In Section~\ref{sec:results}, we extensively discuss the performance of different connectors across various tasks. To consider efficiency and effectiveness simultaneously during the connector selection phase, we present the training times for models under different connectors in Table~\ref{tab:performance_metrics}. It was observed that with increasing image resolution, the training costs for feature-compressing connectors change only slightly, whereas those for feature-preserving connectors significantly increase.

Based on the above evidences, we offer several recommendations for choosing connectors:
\begin{enumerate}
    \item At an image resolution of 224, using a two-layer MLP is advisable as it significantly outperforms other connectors across the three tasks while maintaining a acceptable computational resource demand.
    \item At an image resolution of 336, if the focus is on coarse-grained perception and reasoning tasks, the C-Abstractor and average pooling are recommended for their balance between efficiency and effectiveness. If fine-grained perception tasks are a priority, the two-layer MLP may be more suitable.
    \item At an image resolution of 448, the token count for the two-layer MLP reaches 1024, which leads to excessive consumption of computational resources. Under these circumstances, C-Abstractor and average pooling 144tks emerge as more optimal choices. Specifically, the C-Abstractor reduces the training time by 80\% in the pre-training stage and 51\% in the fine-tuning stage compared to the two-layer MLP. This drastic reduction in training time not only makes the C-Abstractor and average pooling connectors more efficient but also significantly lowers the computational cost, making them highly suitable for scenarios with limited resources. The substantial decrease in training time at this high resolution highlights the importance of choosing the right connector to balance performance and resource usage.
\end{enumerate}

These guidelines aim to assist in selecting the most appropriate connector, aligning with specific task requirements and resource availability.

\begin{table}[ht]
    \centering
    \resizebox{\columnwidth}{!}{
    \begin{tabular}{ccccc}
        \toprule
        \textbf{Resolution} & \textbf{Connectors} & \textbf{Tokens} & \textbf{Stage} & \textbf{Time (hours)} \\
        \midrule
        \multirow{4}{*}{224} & Two-layer MLP & 256 & 1 & 2.0 \\
                             &               & 256 & 2 & 11.3 \\
                             & C-Abstractor  & 144 & 1 & 0.8 (\textcolor{teal}{$\downarrow$60\%}) \\
                             &               & 144 & 2 & 8.0 (\textcolor{teal}{$\downarrow$29\%}) \\
        \midrule
        \multirow{4}{*}{336} & Two-layer MLP & 576 & 1 & 3.6 \\
                             &               & 576 & 2 & 11.9 \\
                             & C-Abstractor  & 144 & 1 & 1.2 (\textcolor{teal}{$\downarrow$67\%}) \\
                             &               & 144 & 2 & 8.0 (\textcolor{teal}{$\downarrow$33\%})\\
        \midrule
        \multirow{4}{*}{448} & Two-layer MLP & 1024 & 1 & 6.5 \\
                             &               & 1024 & 2 & 16.5 \\
                             & C-Abstractor  & 144  & 1 & 1.3 (\textcolor{teal}{$\downarrow$80\%}) \\
                             &               & 144  & 2 & 8.1 (\textcolor{teal}{$\downarrow$51\%}) \\
        \bottomrule
    \end{tabular}
    }
    \caption{Training time for different connectors. Stage 1 refers to the pre-training stage, and Stage 2 refers to the fine-tuning stage. All training is conducted in an environment with 8 A800 GPUs.}
    \label{tab:performance_metrics}
\end{table}

\section{Conclusion}

In this paper, we conduct extensive experiments to evaluate commonly used connectors in MLLMs. Our findings indicate that although feature-preserving connectors generally offer the best performance, their advantage over feature-compressing connectors diminishes as resolution increases, while their computational costs rise exponentially. Among feature-compressing connectors, average pooling and C-Abstractor outperform Q-Former, consistently delivering better results across all resolutions and task granularities. Our results clearly demonstrate that the choice of connector depends on the resolution, task granularity, and computational budget. Based on these findings, we offer guidance on selecting connectors to balance both effectiveness and efficiency.

\section*{Limitations}

In alignment with the base configuration of LLaVA-1.5, our approach involves using positional encoding interpolation to scale images from 336x336 to 448x448, rather than employing a visual encoder that natively supports the 448x448 resolution. This method may lead to suboptimal results. Another limitation is that our training data also comes from LLaVA-1.5, resulting in a relatively small total number of training samples, only 1.23 M. In contrast, the InstructBLIP \citep{dai2024instructblip} with use Q-Former as connector has 130.2 M training samples. This significant difference in the number of training samples might render our conclusions inapplicable in scenarios with a large volume of training data. In the future, we could explore this discrepancy using a larger set of training samples.

\section*{Acknowledgement}
We sincerely thank the reviewers of this work for their constructive and insightful feedback. We thank Xingluan (AI Cloud computing service), EIT and IDT High Performance Computing Center for providing computational resources for this project.

\bibliography{reference}

\newpage

\newpage

\appendix

\section*{Appendix}
We provide some additional information as supplementary material. This appendix is divided into three sections:

\begin{itemize}
    \item Details of sub-task reclassification for MME versus SEED-Bench are presented in Appendix~\ref{sec:reclass};
    \item Experimental details are presented in Appendix~\ref{sec:exp_details};
    \item Additional results are presented in Appendix~\ref{sec:more_re};
\end{itemize}

\section{Sub-Task Reclassification}
\label{sec:reclass}
The parent tasks for sub-tasks of SEED-Bench and MME before and after reclassification are shown in Table~\ref{tab:reclassification_seedbench}, Table~\ref{tab:reclassification_mme} respectively. Table~\ref{tab:all_parenttask} shows all the sub-tasks of MMBench, MME, and SEED-Bench and their reclassified parent tasks.

\begin{table*}[ht]
\centering
\resizebox{\dimexpr 0.95\textwidth}{!}{   
\begin{tabular}{|>{\centering\arraybackslash}m{5cm}|>{\centering\arraybackslash}m{5cm}|>{\centering\arraybackslash}m{5cm}|}
\hline
\textbf{Original Parent Task} & \textbf{Original Sub-Task} & \textbf{New Parent Task} \\ \hline
\multirow{9}{=}{\centering Spatial Understanding} & Scene Understanding & Coarse-grained Perception \\ \cline{2-3} 
 & Instance Identity & Fine-grained Perception \\ \cline{2-3} 
 & Instance Attribute & Fine-grained Perception \\ \cline{2-3} 
 & Instance Location & Fine-grained Perception \\ \cline{2-3} 
 & Instance Counting & Fine-grained Perception \\ \cline{2-3} 
 & Spatial Relation & Fine-grained Perception \\ \cline{2-3} 
 & Instance Interaction & Fine-grained Perception \\ \cline{2-3} 
 & Visual Reasoning & Fine-grained Perception \\ \cline{2-3} 
 & Text Recognition & Fine-grained Perception \\ \hline
\end{tabular}}
\caption{Sub-Task Reclassification for SEED-Bench.}
\label{tab:reclassification_seedbench}
\end{table*}

\begin{table*}[ht]
\centering
\resizebox{\dimexpr 0.95\textwidth}{!}{ 
\begin{tabular}{|>{\centering\arraybackslash}m{5cm}|>{\centering\arraybackslash}m{5cm}|>{\centering\arraybackslash}m{5cm}|}
\hline
\textbf{Original Parent Task} & \textbf{Original Sub-Task} & \textbf{New Parent Task} \\ \hline
\multirow{4}{=}{\centering Perception (Coarse-grained)} & Existence & Fine-grained Perception \\ \cline{2-3} 
 & Count & Fine-grained Perception \\ \cline{2-3} 
 & Position & Fine-grained Perception \\ \cline{2-3} 
 & Color & Fine-grained Perception \\ \hline
\multirow{5}{=}{\centering Perception (Fine-grained)} & Poster & Coarse-grained Perception \\ \cline{2-3} 
 & Celebrity & Fine-grained Perception \\ \cline{2-3} 
 & Scene & Coarse-grained Perception \\ \cline{2-3} 
 & Landmark & Coarse-grained Perception \\ \cline{2-3} 
 & Artwork & Coarse-grained Perception \\ \hline
\multirow{1}{=}{\centering Perception (OCR)} & OCR & Fine-grained Perception \\ \hline
\end{tabular}}
\caption{Sub-Task Reclassification for MME.}
\label{tab:reclassification_mme}
\end{table*}

\begin{table*}[ht]
\centering
\begin{tabular}{|>{\centering\arraybackslash}m{4cm}|>{\centering\arraybackslash}m{4cm}|>{\centering\arraybackslash}m{6cm}|}
\hline
\textbf{Tasks} & \textbf{From} & \textbf{Sub-Tasks} \\ \hline
\multirow{11}{*}{\textbf{Coarse}} & \multirow{5}{*}{MMBench} & Image Quality \\ \cline{3-3} 
 &  & Image Topic \\ \cline{3-3} 
 &  & Image Emotion \\ \cline{3-3} 
 &  & Image Scene \\ \cline{3-3} 
 &  & Image Style \\ \cline{2-3} 
 & \multirow{4}{*}{MME} & Artwork \\ \cline{3-3} 
 &  & Landmark \\ \cline{3-3} 
 &  & Posters \\ \cline{3-3} 
 &  & Scene \\ \cline{2-3} 
 & SEED-Bench & Scene Understanding \\ \hline

\multirow{14}{*}{\textbf{Fine-grained}} & \multirow{7}{*}{MMBench} & OCR \\ \cline{3-3} 
 &  & Celebrity Recognition \\ \cline{3-3} 
 &  & Object Localization \\ \cline{3-3} 
 &  & Attribute Recognition \\ \cline{3-3} 
 &  & Action Recognition \\ \cline{3-3} 
 &  & Attribute Comparison \\ \cline{3-3} 
 &  & Spatial Relationship \\ \cline{2-3} 
 & \multirow{6}{*}{MME} & OCR \\ \cline{3-3} 
 &  & Celebrity \\ \cline{3-3} 
 &  & Color \\ \cline{3-3} 
 &  & Count \\ \cline{3-3} 
 &  & Existence \\ \cline{3-3} 
 &  & Position \\ \cline{2-3} 
 & \multirow{6}{*}{SEED-Bench} & Instance Identity \\ \cline{3-3} 
 &  & Instance Attribute \\ \cline{3-3} 
 &  & Instance Location \\ \cline{3-3} 
 &  & Instance Counting \\ \cline{3-3} 
 &  & Spatial Relationship \\ \cline{3-3} 
 &  & Instance Interaction \\ \hline

\multirow{13}{*}{\textbf{Reasoning}} & \multirow{8}{*}{MMBench} & Function Reasoning \\ \cline{3-3} 
 &  & Identity Reasoning \\ \cline{3-3} 
 &  & Physical Property Reasoning \\ \cline{3-3} 
 &  & Future Prediction \\ \cline{3-3} 
 &  & Image-Text Understanding \\ \cline{3-3} 
 &  & Nature Relation \\ \cline{3-3} 
 &  & Physical Relation \\ \cline{3-3} 
 &  & Social Relation \\ \cline{2-3} 
 & \multirow{4}{*}{MME} & Code Reasoning \\ \cline{3-3} 
 &  & Commonsense Reasoning \\ \cline{3-3} 
 &  & Numerical Calculation \\ \cline{3-3} 
 &  & Text Translation \\ \cline{2-3} 
 & SEED-Bench & Visual Reasoning \\ \hline
\end{tabular}
\caption{Sub-Tasks of MMBench, MME, and SEED-Bench and Their Reclassified Parent Tasks}
\label{tab:all_parenttask}
\end{table*}

\section{Experimental Details}
\label{sec:exp_details}
In this section, we further present the experimental settings, specifically the detailed list of training configurations for different connectors, as shown in Table~\ref{tab:training_config}.

\begin{table*}[htbp]
    \centering
    \begin{tabular}{cccc}
        \toprule
        \textbf{Connector Type} & \textbf{Subclass} & \textbf{Resolution} & \textbf{Token Number} \\
        \midrule
        \multirow{2}{*}{Feature-Preserving} & Linear & \multirow{2}{*}{224, 336, 448} & - \\
        & Nonlinear & & - \\
        \cmidrule(lr){1-4}
        \multirow{3}{*}{Feature-Compressing} & Average Pooling & \multirow{3}{*}{224, 336, 448} & 64, 144 \\
        & Attention Pooling & & 64, 144 \\
        & Convolutional Mapping & & 64, 144 \\
        \bottomrule
    \end{tabular}
    \caption{Connector Configurations.}
    \label{tab:training_config}
\end{table*}

\begin{figure*}[htbp] 
\includegraphics[width=\textwidth]{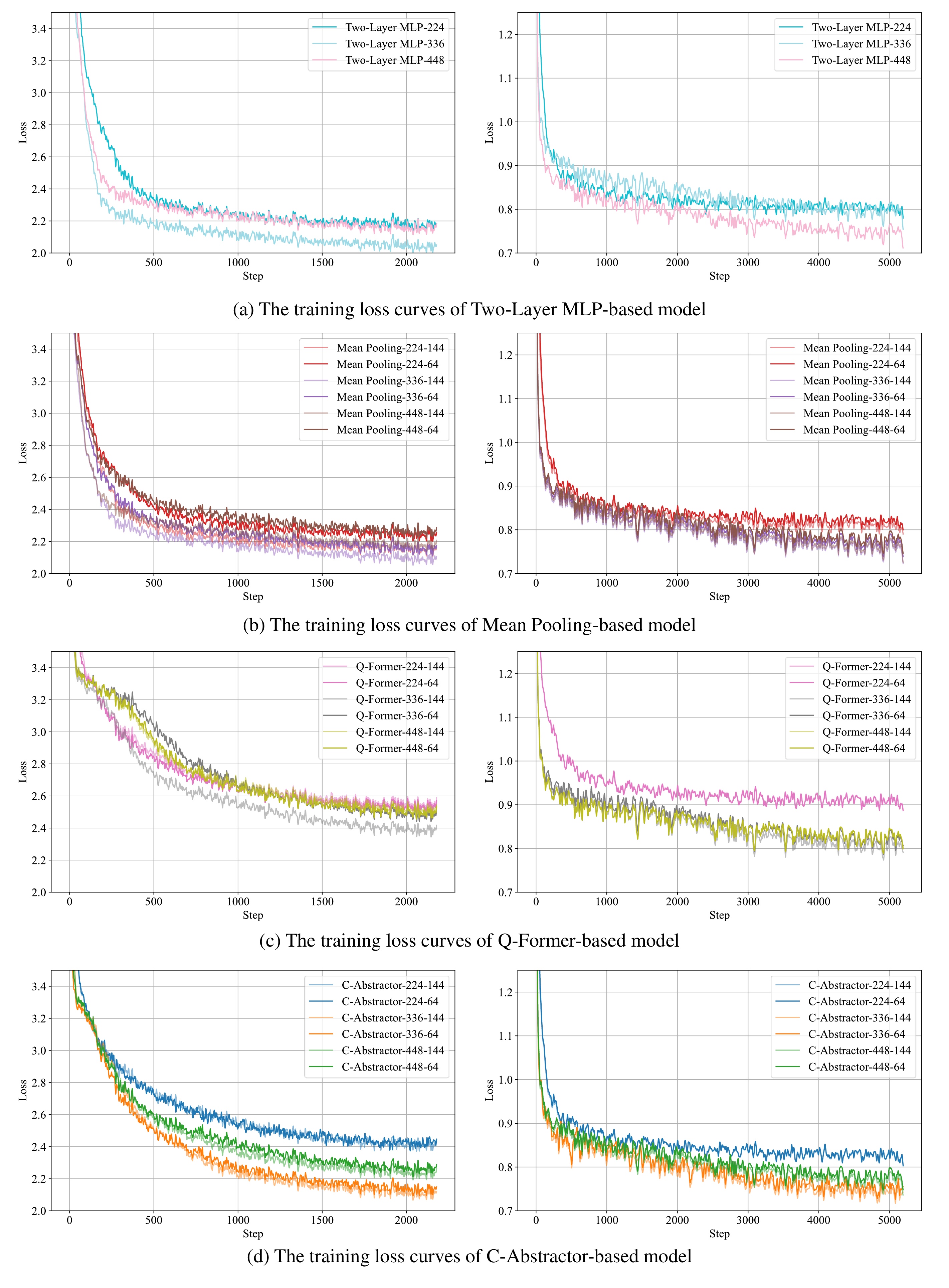}

\caption{Loss curves of different connectors during the pretrain and finetune stages. The left plot shows the training loss during the pretrain stage, and the right plot shows the loss during the finetune stage. The legend in the upper right corner follows the format: connector class-image size-token number (e.g., C-Abstractor-224-144, where the connector is C-Abstractor, the image resolution is 224×224, and the number of tokens is 144). It can be observed that for feature-compressing connectors, the number of tokens has little effect on the loss, while image resolution significantly impacts the model.}
\label{fig:loss_compare}
\end{figure*}

\section{Additional Results}
\label{sec:more_re}
This section provides additional experimental results and analyses to supplement the findings presented in the main text. We include detailed evaluations of loss curves and comprehensive results from additional benchmarks to provide a more thorough understanding of the performance of different connectors.

\subsection{Loss curves for different connectors} 
\label{sec:loss_curves}
Loss curves of the model with different connectors are shown in Figure~\ref{fig:loss_compare}. The loss curves provide several insights that corroborate the findings in the main text: 1. For feature-compressing connectors, the difficulty of convergence increases with the number of parameters and complexity, following the trend: Q-Former > C-Abstractor > Average pooling. 2. When comparing different compressed token numbers, their convergence curves are very similar, with 144 tokens performing slightly better than 64 tokens. 3. Among the image resolutions of 224, 336, and 448, the resolution of 336 often shows significant improvement over 224, but the difference between 448 and 336 is not pronounced, with each resolution occasionally outperforming the other in different scenarios.

\subsection{Evaluation results on more benchmarks}
\label{sec:all_metrices}
To achieve a more extensive and comprehensive comparison, aligning with other works, we conduct experiments on 9 additional benchmarks without sub-task information. These benchmarks include TextVQA \citep{singh2019towards}, POPE \citep{li2023evaluating}, VQAv2 \citep{goyal2017making}, ScienceQA \citep{lu2022learn}, GQA \citep{hudson2019gqa}, RefCOCO, RefCOCO+, RefCOCOg \citep{kazemzadeh2014referitgame}, and VizWiz \citep{gurari2018vizwiz}. The results are shown in Table~\ref{tab:all_metrices}. The metrics used are Exact Match for TextVQA, F1-Score for POPE, Accuracy for VQAv2, GQA, ScienceQA, MMBench, and VizWiz, and CIder for RefCOCO. For SEED-Bench, Accuracy is calculated only on image data.

\begin{table*}[t]
\centering

\resizebox{\textwidth}{!}{
\begin{tabular}{cccccccccccccc}

\toprule
\textbf{Resolution} & \textbf{Projectors} & \textbf{TextVQA} & \textbf{POPE} & \textbf{VQAv2} & \textbf{ScienceQA} & \textbf{GQA} & \textbf{MME\textsuperscript{P}} & \textbf{MMB} & \textbf{SEED} & \textbf{Refcoco} & \textbf{Refcoco+} & \textbf{Refcocog} & \textbf{VizWiz} \\ 
\midrule
\multirow{8}{*}{244} & Linear & 46.88 & 81.06 & 69.54 & 63.97 & 50.29 & 1227.7 & 50.94 & 53.53 & 15.96 & 15.58 & 46.07 & 53.3 \\ 
& Two-layers MLP & 49.08 & 81.89 & 73.26 & 64.77 & 53.8 & 1309.91 & 55.67 & 58.3 & 23.2 & 23.03 & 52.79 & 55.26 \\ 
& Average pooling 64tks & 46.66 & 80.53 & 70.52 & 64.61 & 52.25 & 1294.79 & 53.78 & 56.3 & 20.09 & 19.43 & 49.14 & 55.63 \\ 
& Average pooling 144tks & 49.61 & 82.03 & 73.17 & 64.18 & 53.75 & 1333.86 & 54.04 & 58.63 & 23.44 & 23.32 & 54.57 & 54.7 \\ 
& Q-Former 64tks & 34.47 & 83.07 & 58.33 & 58.29 & 35.8 & 1050.59 & 36.77 & 46.09 & 8.82 & 8.34 & 32.22 & 24.04 \\ 
& Q-Former 144tks & 33.49 & 76.98 & 57.81 & 60.53 & 28.36 & 1127.66 & 37.2 & 45.8 & 2.66 & 2.49 & 13.41 & 18.06 \\
& C-Abstactor 64tks & 42.29 & 80.15 & 66.55 & 63.1 & 51.39 & 1228.72 & 48.79 & 52.74 & 18.45 & 17.74 & 50.65 & 53.9 \\ 
& C-Abstactor 144tks & 43.65 & 81.19 & 66.46 & 63.57 & 50.89 & 1204.11 & 48.45 & 52.87 & 9 & 8.75 & 38.11 & 52.87 \\ 
\midrule
\multirow{8}{*}{336} & Linear & 56.16 & 86.06 & 78.41 & 70.08 & 62.53 & 1431.73 & 65.38 & 65.74 & 28.64 & 27.62 & 63.03 & 55.58 \\
& Two-layers MLP & 56.37 & 85.63 & 78.65 & 70.97 & 63.24 & 1486.42 & 65.12 & 67.08 & 28.65 & 28.15 & 62.82 & 56.22 \\
& Average pooling 64tks & 50.99 & 84.01 & 74.76 & 69.77 & 59.21 & 1388.58 & 63.05 & 61.49 & 24.73 & 23.77 & 57.97 & 51.84 \\
& Average pooling 144tks & 53.88 & 85.03 & 76.4 & 70.17 & 60.14 & 1457.42 & 63.4 & 63.98 & 28.88 & 28.71 & 60.3 & 53.54 \\
& Q-Former 64tks & 45.34 & 80.97 & 67.48 & 70.41 & 53.82 & 1244.26 & 57.73 & 53.73 & 25.5 & 24.81 & 57.14 & 49.88 \\
& Q-Former 144tks & 46.87 & 76.07 & 67.16 & 70.88 & 53.34 & 1240.74 & 58.59 & 53.09 & 29.6 & 27.99 & 57.28 & 49.52 \\
& C-Abstactor 64tks & 54.34 & 85.21 & 76.77 & 70.05 & 60.38 & 1360 & 63.49 & 62.55 & 25.35 & 24.51 & 61.22 & 54.37 \\
& C-Abstactor 144tks & 54.36 & 85.07 & 76.77 & 71.4 & 61.07 & 1450.46 & 63.57 & 63.42 & 28.76 & 28.6 & 60.71 & 55.13 \\
\midrule
\multirow{8}{*}{448} & Linear & 55.73 & 84.92 & 77.65 & 68.56 & 60.93 & 1458 & 63.74 & 65.03 & 29.98 & 29.44 & 62.17 & 54.41 \\
& Two-layers MLP & 56.32 & 84.55 & 78.97 & 69.42 & 61.28 & 1516.15 & 65.12 & 66.74 & 28.75 & 28.66 & 62.56 & 57.93 \\
& Average pooling 64tks & 50.54 & 82.97 & 75.07 & 70.01 & 59.19 & 1422.03 & 62.29 & 62.19 & 27.98 & 27.89 & 62.01 & 56.76 \\
& Average pooling 144tks & 54.32 & 84.55 & 68.3 & 69.42 & 60.57 & 1474.24 & 63.48 & 64.91 & 27.83 & 27.58 & 64.05 & 53.28 \\
& Q-Former 64tks & 45.41 & 81.72 & 67.26 & 69.37 & 52.28 & 1281.34 & 56.96 & 54.49 & 27.4 & 26.51 & 55.07 & 46.98 \\
& Q-Former 144tks & 44.82 & 81.2 & 66.3 & 69.49 & 52.34 & 1236.01 & 57.82 & 52.86 & 26.39 & 25.71 & 56.07 & 50.88 \\
& C-Abstactor 64tks & 50.44 & 83.02 & 74.65 & 69.72 & 58.3 & 1374.8 & 62.71 & 61.67 & 25.99 & 25.7 & 57.41 & 54.62 \\
& C-Abstactor 144tks & 51.9 & 84.51 & 75.71 & 69.7 & 59.55 & 1433.33 & 63.23 & 62.49 & 27.98 & 27.75 & 59.85 & 55.3 \\
\bottomrule
\end{tabular}
}
\caption{Performance of different connectors across 12 datasets at resolutions of 224, 336, and 448. Here, MME\textsuperscript{P} denotes the MME-Perception tasks, MMB stands for MMBench, and SEED indicates SEED-Bench.}
\label{tab:all_metrices}
\end{table*}

\end{document}